\documentclass[letterpaper, 10 pt, conference]{ieeeconf}  % Comment this line out if you need a4paper

\IEEEoverridecommandlockouts                              % This command is only needed if 
\usepackage{graphicx}
\usepackage{booktabs}
\usepackage {tablefootnote}
\usepackage{amssymb} % for \checkmark

\usepackage{amsmath}
\usepackage{algorithmic}
\usepackage{algorithm}
\usepackage{siunitx}
\usepackage{multirow}
\usepackage{url}
\usepackage{cite}
\usepackage[hang,small,bf]{caption}
\usepackage[subrefformat=parens]{subcaption}
\usepackage{placeins}

\title{\LARGE \bf MRReP: Mixed Reality-based Hand-drawn Reference Path Editing Interface for Mobile Robot Navigation}

\author{Takumi Taki$^{1\dag}$, Masato Kobayashi$^{1,2,3\dag*}$, Yuki Uranishi$^{1,2}$ % <-
\thanks{
${\dag}$ Equal Contribution,
$^{1}$ Graduate School of Information Science and Technology, The University of Osaka, $^{2}$ D3 Center, The University of Osaka, $^{3}$ Graduate School of Maritime Sciences, Kobe University, * corresponding author: kobayashi.masato.cmc@osaka-u.ac.jp}
}

\begin{document}

\maketitle
\thispagestyle{empty}
\pagestyle{empty}

%%%%%%%%%%%%%%%%%%%%%%%%%%%%%%%%%%%%%%%%%%%%%%%%%%%%%%%%%%%%%%%%%%%%%%%%%%%%%%%%
\begin{abstract}
Autonomous mobile robots operating in human-shared indoor environments often require paths that reflect human spatial intentions, such as avoiding interference with pedestrian flow or maintaining comfortable clearance. However, conventional path planners primarily optimize geometric costs and provide limited support for explicit route specification by human operators. This paper presents MRReP, a Mixed Reality-based interface that enables users to draw a Hand-drawn Reference Path (HRP) directly on the physical floor using hand gestures. The drawn HRP is integrated into the robot navigation stack through a custom Hand-drawn Reference Path Planner, which converts the user-specified point sequence into a global path for autonomous navigation. We evaluated MRReP in a within-subject experiment against a conventional 2D baseline interface. The results demonstrated that MRReP enhanced path specification accuracy, usability, and perceived workload, while enabling more stable path specification in the physical environment. These findings suggest that direct path specification in MR is an effective approach for incorporating human spatial intention into mobile robot navigation. Additional material is available at \url{https://mertcookimg.github.io/mrrep/}
\end{abstract}

\section{Introduction}
Autonomous mobile robots are increasingly being deployed in indoor environments such as warehouses, hospitals, and offices~\cite{zghair2021one, eriksen2023understanding, herath2023robots}. In these settings, navigation is commonly performed on occupancy grid maps using global planners such as A*, which generate paths mainly by optimizing geometric criteria such as path length and obstacle-related costs. While such planners are effective for computing collision-free paths, they provide limited support for cases in which a human operator wishes to explicitly specify how the robot should move through the environment.

In practical operation, there are many situations in which an operator has a preferred path in mind. For example, the operator wants the robot to pass through a wider area, avoid a locally undesirable region, or follow a particular side of a corridor. These preferences are often spatial and geometric in nature: they concern the shape and placement of the path itself rather than only the final destination. Although such intentions may sometimes relate to social or operational considerations, the more fundamental issue is that conventional navigation interfaces provide limited means for directly specifying an intended path. As a result, operators often have to rely on indirect methods such as cost tuning, repeated replanning, or discrete goal and waypoint input, which may not faithfully capture the desired path.

Several approaches have been explored to bridge this gap. Conventional 2D interfaces allow users to edit maps or draw paths, but they require mental translation between a 2D map and the 3D physical environment~\cite{sidaoui2018human, skubic2007team, sakamoto2009sketchandrun}. This translation can make it difficult to specify a path that precisely matches the operator's spatial intention in the real workspace. Natural-language-based approaches can provide high-level guidance, but they are often too ambiguous for specifying detailed geometric paths. Recent mixed reality (MR) interfaces have enabled more intuitive spatial interaction in the physical environment. For example, MR-based systems have enabled users to specify goals or define no-go regions directly in the environment~\cite{taki2025mrhad}. However, these approaches mainly support discrete destination specification or negative constraints, rather than direct specification of a continuous path.

\begin{figure}[t]
    \centering
    \includegraphics[keepaspectratio, width=0.95\linewidth]{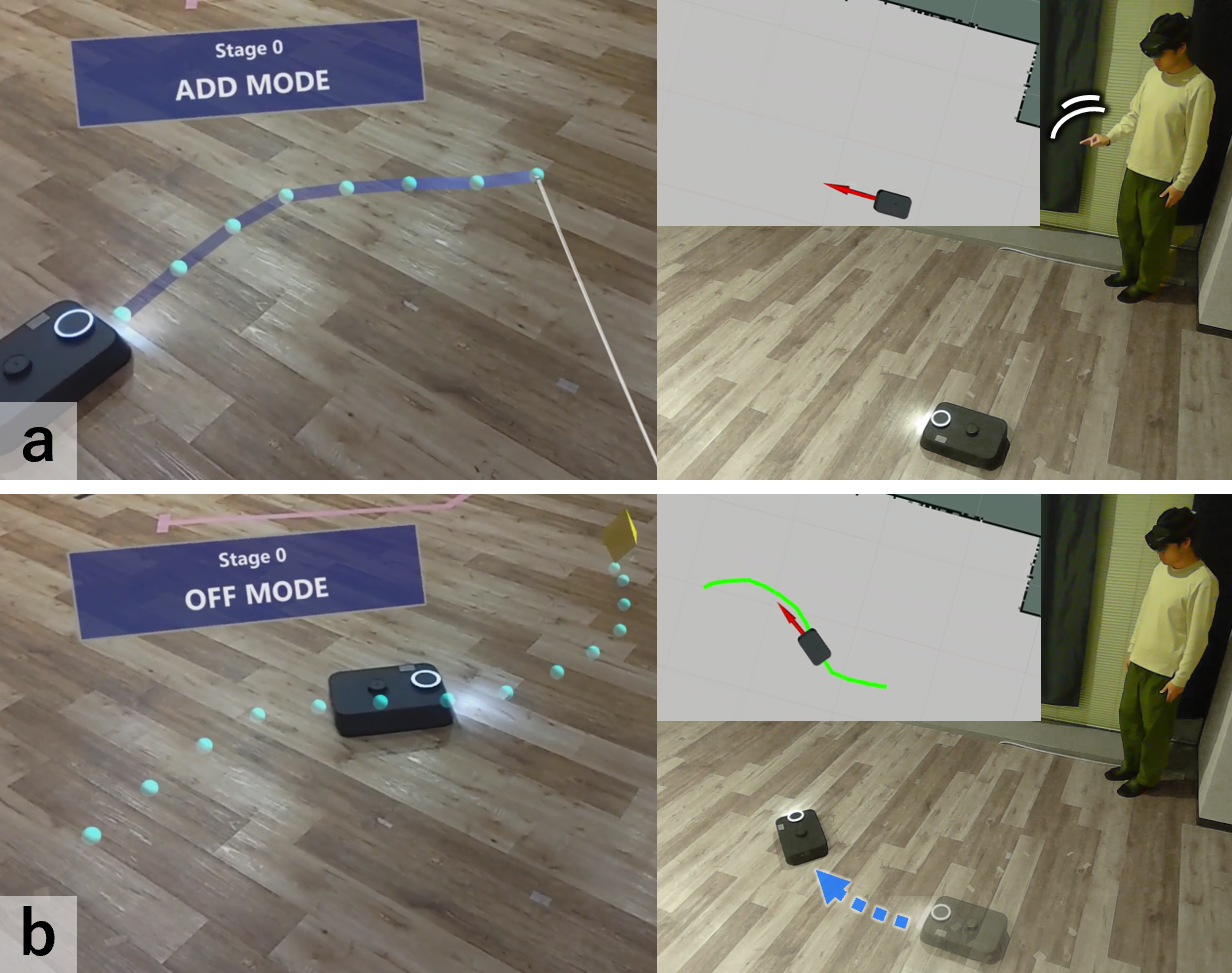}
    \caption{Overview of MRReP. (a) A user draws a Hand-drawn Reference Path (HRP) directly on the physical floor using hand gestures and an MR-HMD. (b) The HRP is sent to the robot navigation system, enabling autonomous navigation along the specified path.}
    \label{fig:teaser}
\end{figure}

To address this limitation, this paper presents MRReP, a Mixed Reality-based interface for specifying robot navigation paths directly in the physical environment through hand gestures. As shown in Fig.~\ref{fig:teaser}, the user draws a Hand-drawn Reference Path (HRP) directly on the physical floor while observing the real environment through an MR head-mounted display. By allowing users to specify the path in the same physical space in which the robot moves, MRReP reduces the gap between screen-based editing and real-world spatial intention. In addition, we develop a custom Hand-drawn Reference Path Planner that converts the drawn HRP into a global path within the robot navigation stack, enabling the robot to navigate along the user-specified path.

The contributions of this paper are as follows:
\begin{itemize}
    \item We propose MRReP, an MR-based interface that enables users to draw robot navigation paths directly in the physical environment using hand gestures.
    \item We develop a custom Hand-drawn Reference Path Planner that converts the drawn point sequence into a global path in the ROS~2 navigation stack.
    \item We conduct a within-subject experiment showing that MRReP enables more faithful and intuitive path specification directly in the physical environment than a conventional 2D interface, improving path specification accuracy and subjective usability while reducing perceived workload.
\end{itemize}

The remainder of this paper is organized as follows. Section~\ref{sec:RW} reviews related work. Sections~\ref{sec:SD} and~\ref{sec:SI} describe the system design and implementation. Section~\ref{sec:EXP} presents the experiment and results. Section~\ref{sec:CON} concludes the paper.

\section{Related Work}\label{sec:RW}

\subsection{Robot Navigation Interfaces}
Robot navigation interfaces have traditionally been built around 2D graphical user interfaces (GUIs), such as RViz~\cite{kam2015rviz}. These interfaces are effective for specifying goals and waypoints on a map, but path generation itself is delegated to the navigation stack. As a result, they provide limited support for directly reflecting a user-intended geometric route.

More recently, high-level instruction methods based on natural language and visual information have been proposed. These include approaches that generate cost maps from language, incorporate semantic information to improve navigation safety, and perform navigation directly from RGB observations~\cite{shah2023lm, yu2025co, bao2025path, omer2024semantic, sridhar2024nomad}.

Immersive spatial interfaces using VR/MR have also been explored for more intuitive robot interaction. Prior work includes path-planning support and visualization~\cite{wu2020mixed, yu2022mixed, hoang2022arviz}, goal and waypoint specification~\cite{iglesius2025mrnab, fang2025mixed, gu2022arpointclick, iglesius2026mrpos, baker2020towards}, constraint-region editing~\cite{taki2025mrhad}, and navigation assistance in VR environments~\cite{solanes2022virtual}. However, most of these approaches focus on specifying destinations or constraints rather than directly providing a continuous path in the physical environment.

In contrast, MRReP enables users to directly draw a Hand-drawn Reference Path (HRP) on the physical floor in an MR environment, and the resulting point sequence is used as the robot's global path. Thus, MRReP is positioned as a navigation interface for explicitly specifying \textit{how} the robot should move, whereas existing approaches focus primarily on specifying \textit{where} it should go or \textit{what} it should avoid.

\begin{table}[t]
    \centering
    \caption{Comparison of selected path-specification interfaces.}
    \label{tab:related_works}
    \begin{tabular}{lccccc}
        \toprule
         & C1 & C2 & C3 & C4 & \textbf{MRReP} (Ours) \\
        \midrule
        Path specification accuracy & $\times$ & $\times$ & $\times$ & $\times$ & \textbf{$\checkmark$}\\
        Path following accuracy & $\times$ & $\checkmark$ & $\checkmark$ & $\checkmark$ & \textbf{$\checkmark$}\\
        Spatial intuitiveness & $\times$ & $\times$ & $\times$ & $\times$ & $\checkmark$\\
        No fixed sensor required & $\checkmark$ & $-$ & $\times$ & $\checkmark$ & \textbf{$\checkmark$}\\
        \bottomrule
    \end{tabular}

    C1: Sketching rough maps and paths ~\cite{boniardi2016sketch, tan2025mobile}, C2: Sketching on 2D maps/images and navigation exclusively in a simulation environment~\cite{hwang2003fingertip}, C3: Sketching on 2D maps/images and sensing with fixed cameras~\cite{sakamoto2009sketchandrun, hou2020novel, frank2015pathbending, zu2024language}, C4: Sketching on 2D maps/images and sensing without fixed cameras~\cite{skubic2007team, choi2025canvas}
\raggedright    
\end{table}

\subsection{Path Editing Interfaces}
One line of work on explicit path specification uses rough or hand-drawn maps as input. For example, prior work has allowed users to draw simplified maps together with intended routes and use them for navigation~\cite{boniardi2016sketch, tan2025mobile}. While these approaches do not require a precise prebuilt map, they are less suitable for fine path adjustment based on spatial relationships in the real environment.

Another line of work allows users to draw paths on 2D maps or images and uses the resulting trajectories for robot path following. Examples include methods that convert sketches or touch input into executable trajectories~\cite{skubic2003sketch, skubic2007team, hwang2003fingertip, sakamoto2009sketchandrun} and methods that refine user-drawn trajectories under map constraints to generate feasible paths~\cite{hou2020novel, frank2015pathbending}. Although these approaches allow explicit route specification, they rely on 2D media and therefore require users to mentally translate between the real environment and its 2D representation.

More recent work has combined sketch input with language or additional reasoning mechanisms~\cite{zu2024language, choi2025canvas}. Although these methods support more flexible instruction, path specification still depends on 2D representations or indirect inference.

In contrast, MRReP allows users to directly draw a continuous HRP on the physical floor in an MR environment and uses the resulting point sequence as a global path. This reduces the spatial translation burden inherent in 2D-based approaches and more directly reflects the user's spatial intention in robot navigation.

Table~\ref{tab:related_works} further highlights the position of MRReP among prior path specification interfaces. MRReP supports both direct path following and spatially intuitive path specification in the physical environment without requiring fixed external sensors.

\begin{figure}[t]
    \centering
    \includegraphics[keepaspectratio, width=\linewidth]{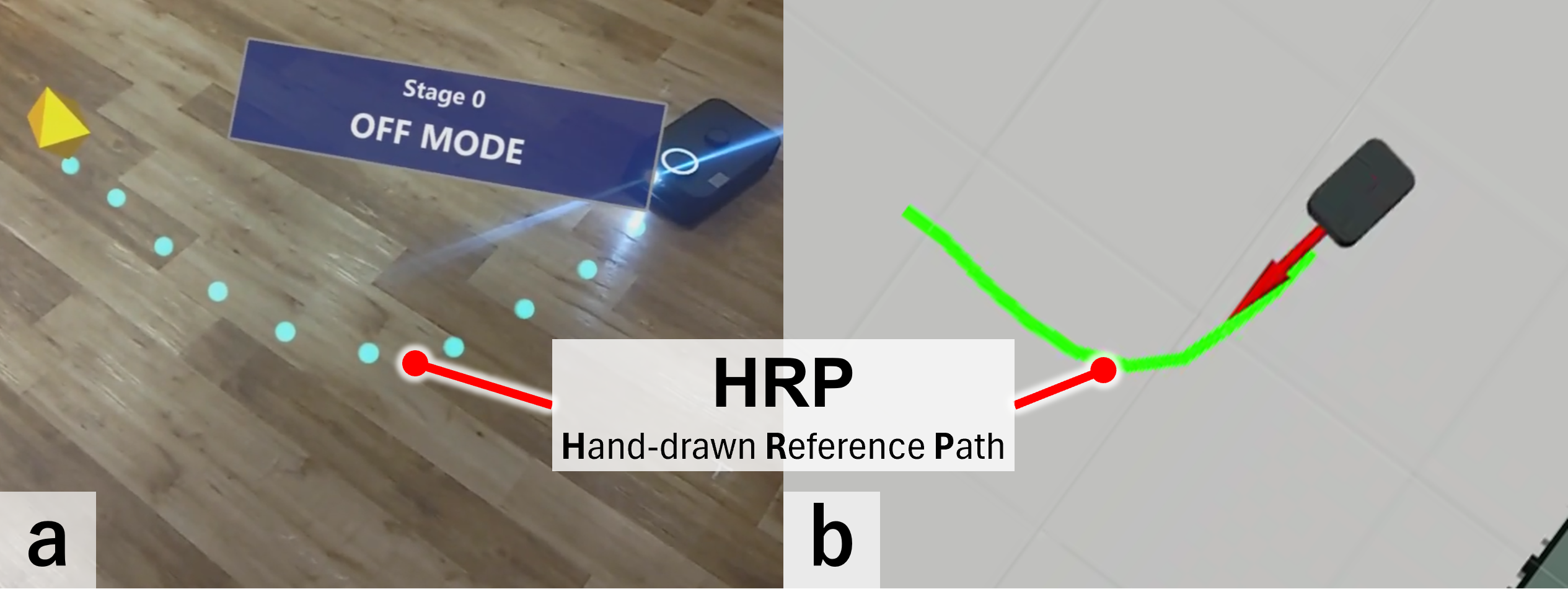}
    \caption{Example of an HRP and its corresponding global path. (a) HRP overlaid on the floor in HoloLens~2; the yellow pin indicates the goal. (b) Generated global path and costmap.}
    \label{fig:hrp_sample}
\end{figure}

\begin{figure*}[t]
    \centering
    \includegraphics[keepaspectratio, width=1\linewidth]{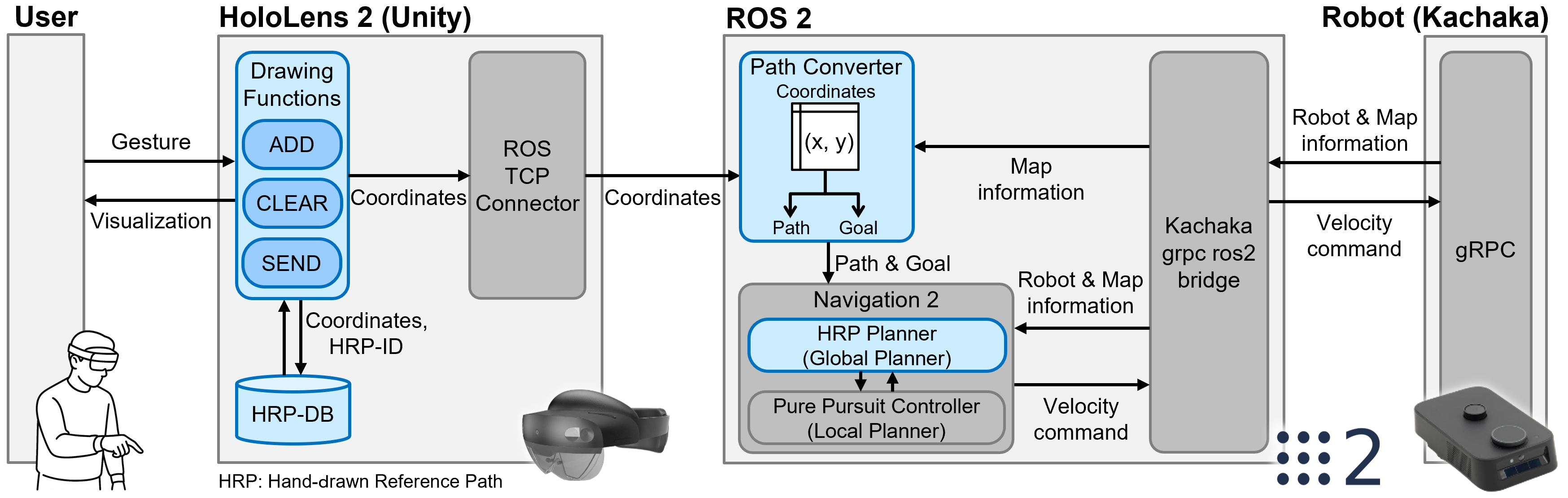}
    \caption{System architecture and data flow. The drawn HRP is stored in the HoloLens~2 database, transmitted to ROS~2 via the SEND function, converted into a global path and goal pose, and then used for robot navigation in Navigation2.}
    \label{fig:system_structure}
\end{figure*}

\section{System Design}\label{sec:SD}

\begin{figure}[t]
    \centering
    \includegraphics[keepaspectratio, width=1\linewidth]{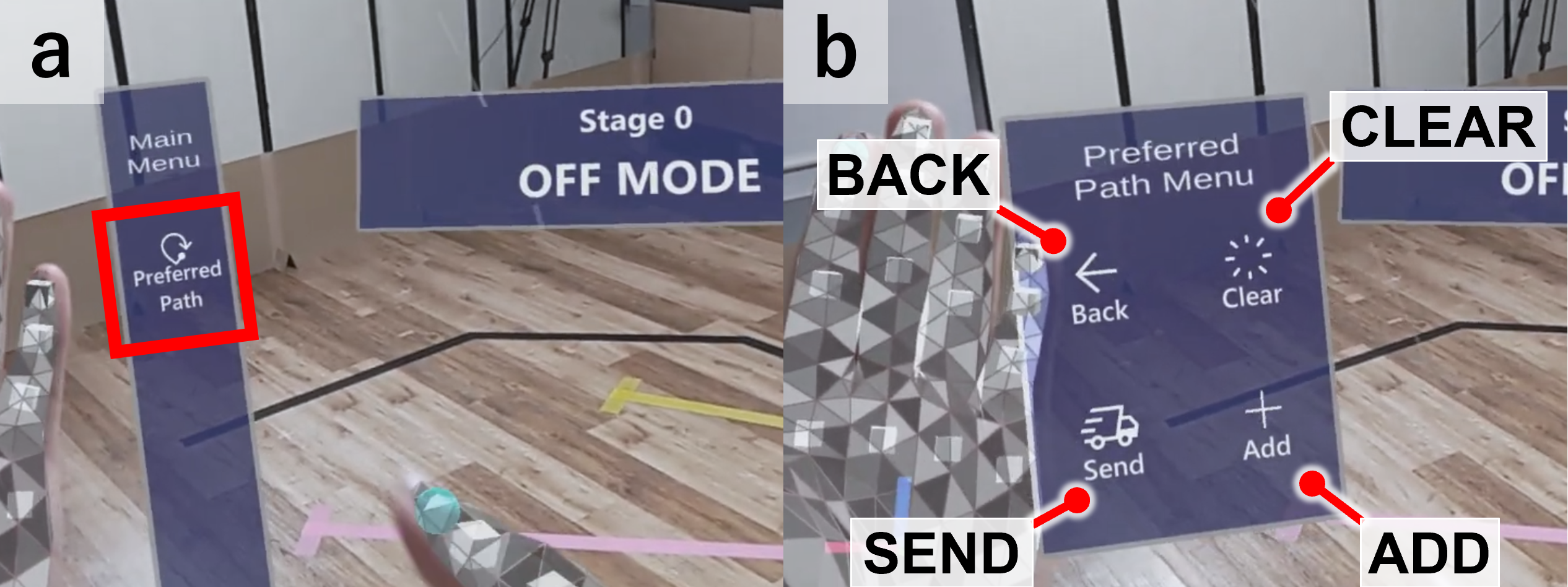}
    \caption{Menu panel. (a) Main menu in the initial state. (b) Path operation menu with selectable operation modes.}
    \label{fig:menu_panel}
\end{figure}

The proposed system allows users wearing a HoloLens~2 to draw a Hand-drawn Reference Path (HRP) directly in the physical environment through hand gestures. The drawn HRP is visualized in the MR view and stored together with its ID and coordinate sequence, enabling the previous state to be restored after reboot. When transmitted to ROS~2, the stored coordinates are converted into a global path for autonomous robot navigation. Fig.~\ref{fig:hrp_sample} shows an example of an HRP in HoloLens~2 and its corresponding global path in ROS~2, and Fig.~\ref{fig:system_structure} illustrates the overall architecture and data flow.

The interface provides three fundamental functions for HRP manipulation: ADD, CLEAR, and SEND. These functions correspond to three active modes, in addition to an OFF mode. Users switch among these modes through the spatial menu panel shown in Fig.~\ref{fig:menu_panel}.

\begin{figure}[t]
    \begin{minipage}[b]{1\linewidth}
        \centering
        \includegraphics[keepaspectratio, width=1\linewidth]{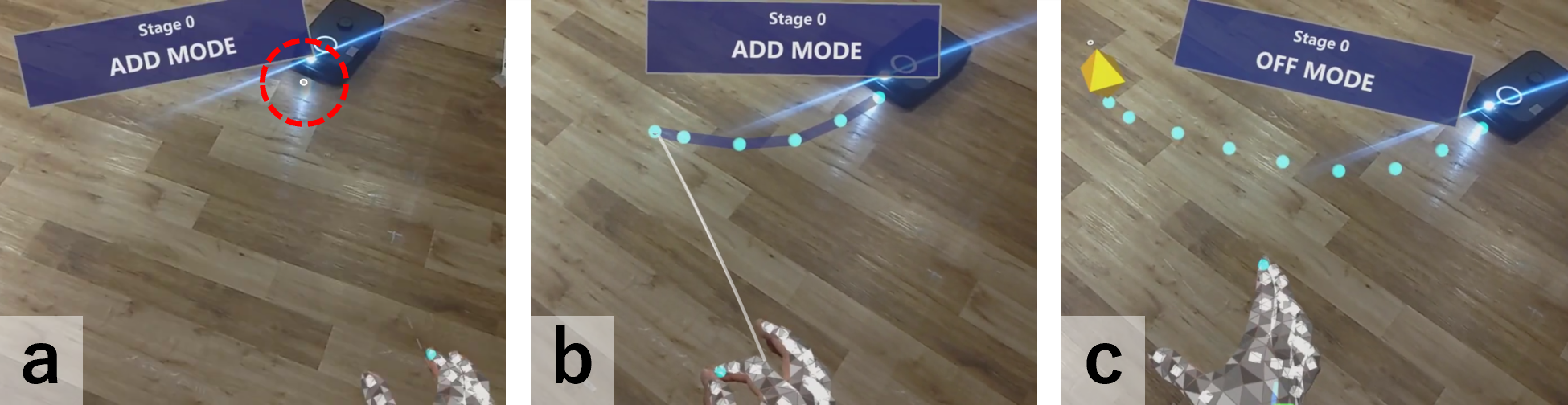}
        \vspace{-5mm}
        \caption{ADD operation. (a) A floor cursor appears when the user extends the hand. (b) Maintaining a pinch gesture draws an HRP as a sequence of waypoints. (c) Releasing the pinch completes the path and places a goal pin at the endpoint.}
        \label{fig:function_add}
    \end{minipage}
\end{figure}

\subsection{ADD Function}\label{sec:SD_function_add}
The ADD function is designed to create a new HRP. In ADD mode, the user generates a piecewise linear path by manipulating a floor cursor through hand gestures. Once the drawing process is completed, the system assigns a unique ID to the HRP and stores its corresponding coordinate sequence in the database. The interaction procedure is illustrated in Fig.~\ref{fig:function_add}.

\begin{figure}[t]
    \begin{minipage}[b]{1\linewidth}
        \centering
        \includegraphics[keepaspectratio, width=1\linewidth]{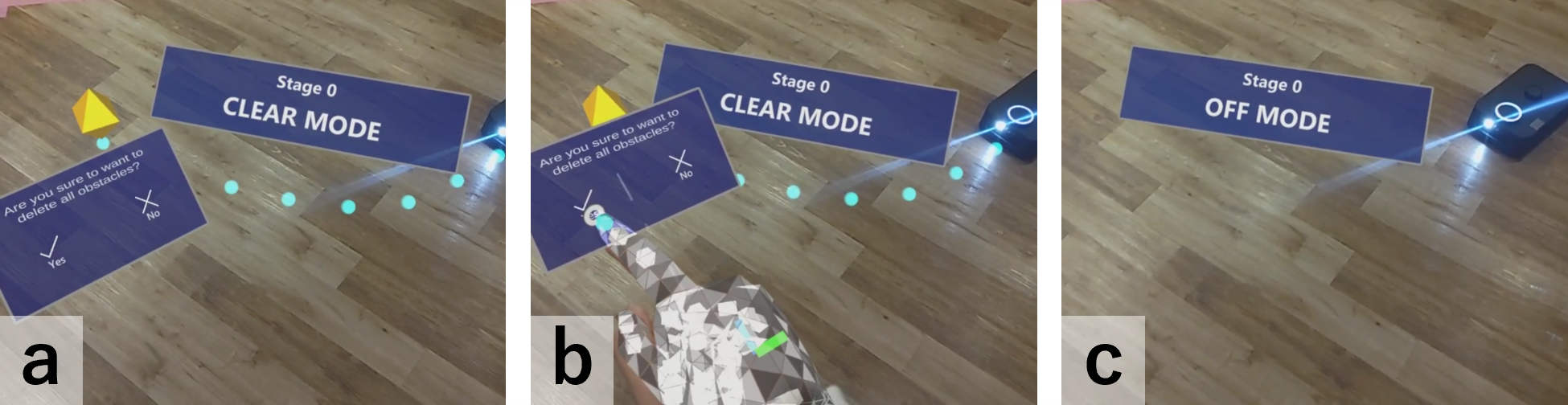}
        \vspace{-4mm}
        \caption{CLEAR operation. (a) A confirmation popup appears. (b) When the user selects ``Yes,'' (c) all HRPs are removed from the scene.}
        \label{fig:function_clear}
    \end{minipage}
\end{figure}

\subsection{CLEAR Function}\label{sec:SD_function_clear}
The CLEAR function deletes all existing HRPs. When CLEAR mode is selected, the system displays a confirmation popup to prevent accidental deletion. If confirmed, all HRP objects are removed from the scene, the corresponding coordinate records are deleted from the database, and the system returns to OFF mode. The procedure is shown in Fig.~\ref{fig:function_clear}.

\begin{figure}[t]
    \begin{minipage}[b]{1\linewidth}
        \centering
        \includegraphics[keepaspectratio, width=1\linewidth]{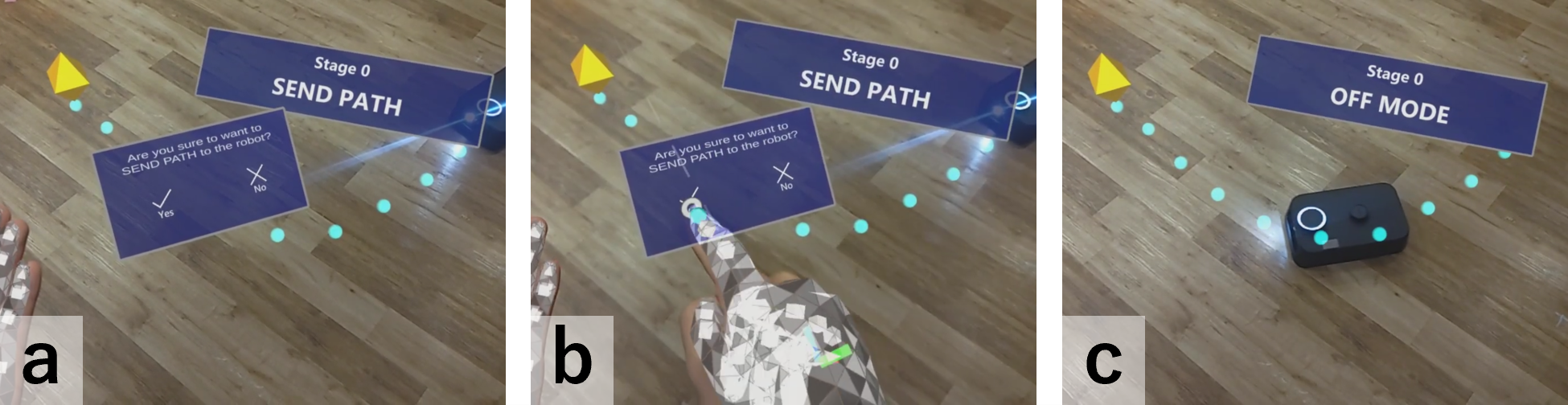}
        \vspace{-4mm}
        \caption{SEND operation. (a) A confirmation popup appears. (b) The stored HRP coordinates are transmitted to ROS~2. (c) The robot starts path-following navigation based on the converted global path.}
        \label{fig:function_send}
    \end{minipage}
\end{figure}

\subsection{SEND Function}\label{sec:SD_function_send_path}

The SEND function transmits the stored HRP data to ROS~2. After confirmation, the coordinate sequence of the HRP stored in the database is published as a ROS~2 message. Furthermore, path following along the HRP is executed using the Nav2 stack in ROS 2. The procedure is shown in Fig.~\ref{fig:function_send}.

\section{System Implementation}\label{sec:SI}

\subsection{Communication Method}
The proposed system uses two communication links: Unity--ROS~2 and ROS~2--Kachaka. Unity and ROS~2 communicate via ROS-TCP-Connector and ROS-TCP-Endpoint. ROS-TCP-Endpoint runs on the ROS~2 side as a TCP server, while Unity uses a ROS-TCP-Connector configured with the IP address and port of the ROS~2 host to exchange topic messages. Communication between ROS~2 and Kachaka is handled through the gRPC-based kachaka-api, which provides robot control and state retrieval functions~\cite{kachakaapi}.

\subsection{Coordinate Alignment}
The coordinate systems of Unity and ROS~2 are aligned through HoloLens~2-based QR code recognition using the Vuforia Engine. A physical QR code is placed at the origin of the ROS~2 coordinate system, and a corresponding virtual QR object is placed at the origin of the Unity space. By matching the detected physical code with the virtual reference, the transformation between the Unity and ROS~2 map coordinates is established.

\subsection{Path Data Management}
To restore the previous state after reboot, the system stores the coordinate sequence of the HRP in a JSON file managed by a Unity script. The record contains an array of coordinates defined relative to the QR-code origin. As outlined in Algorithm~\ref{alg:database}, the stored data are loaded at startup to instantiate the corresponding path object. The JSON file is overwritten whenever the path is added or cleared. During transmission, the stored coordinate sequence is sent to ROS~2.

\begin{algorithm}[tb]
    \small
    \caption{Path Data Management Process}
    \label{alg:database}
    \begin{algorithmic}[1]
    \STATE $database \leftarrow$ LoadJsonFile()
    \STATE InstantiateAllHRPs($database$)
    \WHILE{system is running}
        \IF{$currentFunction$ is ADD}
            \STATE $database \leftarrow$ LoadJsonFile()
            \STATE SaveDatabase($database$, $\mathit{path\_id}$, $coordinates$)
            \STATE OverwriteJsonFile($database$)
        \ELSIF{$currentFunction$ is CLEAR}
            \STATE $database \leftarrow$ GenerateEmptyJsonFile()
            \STATE OverwriteJsonFile($database$)
        \ELSIF{$currentFunction$ is SEND}
            \STATE $database \leftarrow$ LoadJsonFile()
            % \STATE $coordinates \leftarrow$ ConcatenatePathsInIdOrder($database$)
            \STATE $coordinates \leftarrow$ FetchPathCoordinates($database$)
            \STATE SendPathToROS2($coordinates$)
        \ENDIF
    \ENDWHILE
    \end{algorithmic}
\end{algorithm}

\subsection{Addition and Deletion of HRP in Map}\label{subsec:addition_and_clearing}
HRP addition and deletion are implemented through dynamic instantiation and destruction of Unity GameObjects. During drawing, a new waypoint is generated when the distance between the current cursor position and the previous waypoint exceeds a predefined threshold $D_{\mathrm{th}}$. When the pinch gesture is released, a goal pin is placed above the final waypoint, and the resulting coordinate sequence is stored in the database. Conversely, clearing deletes all waypoint and goal-pin objects and removes the corresponding records from the database.

\subsection{Path Processing and Navigation}
\label{subsec:path_processing_and_navigation}
The transmitted HRP point sequence is converted into a global path and used for autonomous robot navigation. For this purpose, we implemented a custom node that converts the received points into a Navigation2-compatible path and a custom global planner that provides the converted path to the navigation stack. The ROS~2-side processing flow is as follows:

\begin{enumerate}
    \item \textbf{Path Data Reception}: The HRP point sequence is received from Unity.
    \item \textbf{Global Path Generation}: A global path is generated from the received points, and an orientation is assigned to each point based on the direction to its adjacent point.
    \item \textbf{Navigation Initialization}: The final point is published as the goal pose to trigger navigation in Navigation2.
    \item \textbf{Path Provision}: The custom global planner retains the latest converted path and returns it when Navigation2 requests a global path.
    \item \textbf{Velocity Command Calculation}: A Pure Pursuit controller~\cite{macenski2023regulated} computes translational and angular velocities from a locally transformed path and a target point selected with a predefined lookahead distance.
\end{enumerate}

\section{Experiment}\label{sec:EXP}

\begin{figure}[t]
    \centering
    \includegraphics[keepaspectratio, width=\linewidth]{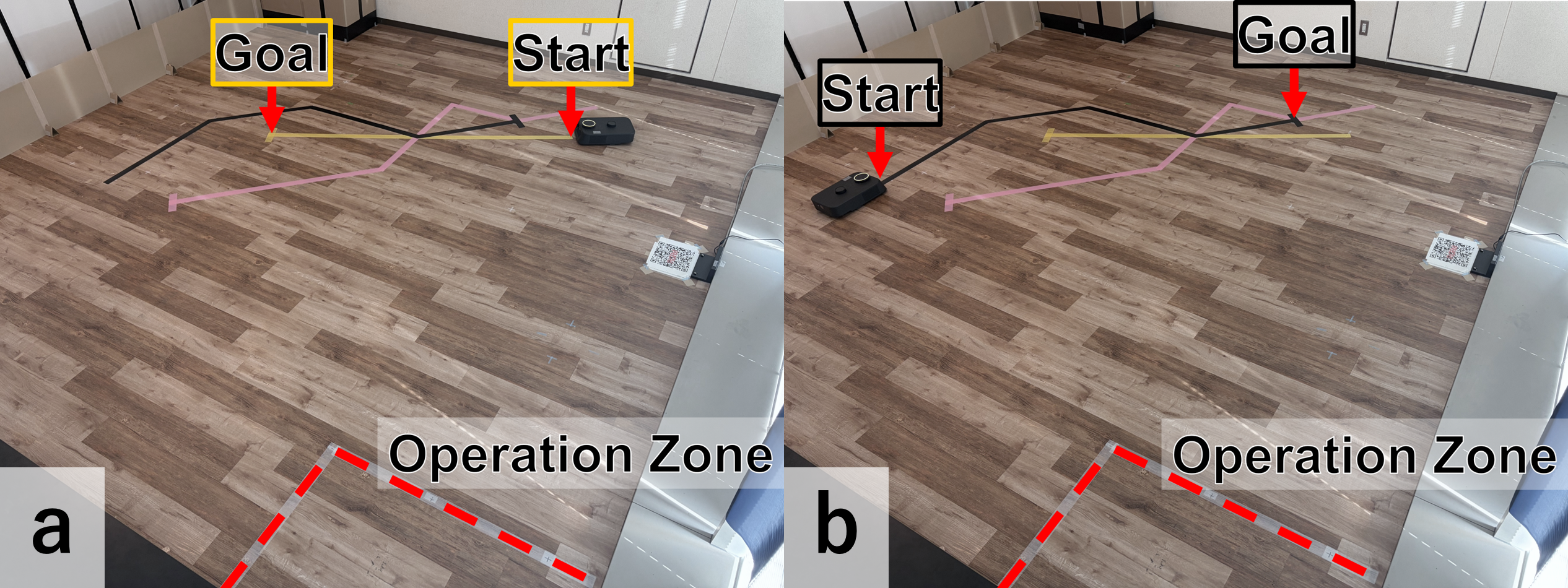}
    \caption{Experimental environment. (a) Stage A: straight path. (b) Stage B: piecewise linear path with multiple 45-degree turns.}
    \label{fig:experiment_environment}
\end{figure}

\begin{figure}[t]
    \centering
    \includegraphics[keepaspectratio, width=\linewidth]{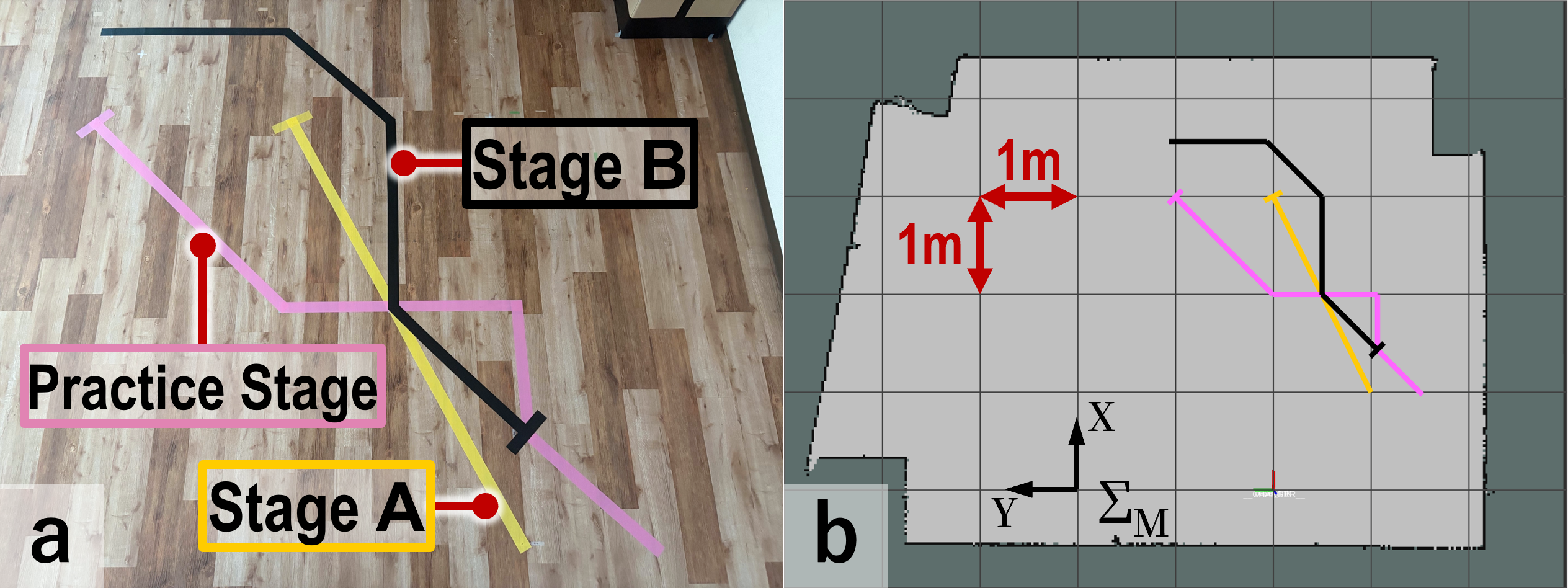}
    \caption{Target paths used in the practice and main tasks. (a) Tape layout in the experimental environment. (b) Corresponding paths on the ROS~2 costmap. T-shaped markers indicate the goal positions.}
    \label{fig:experimental_environment_gt_map}
\end{figure}

\begin{figure}[t]
    \centering
    \includegraphics[keepaspectratio, width=0.9\linewidth]{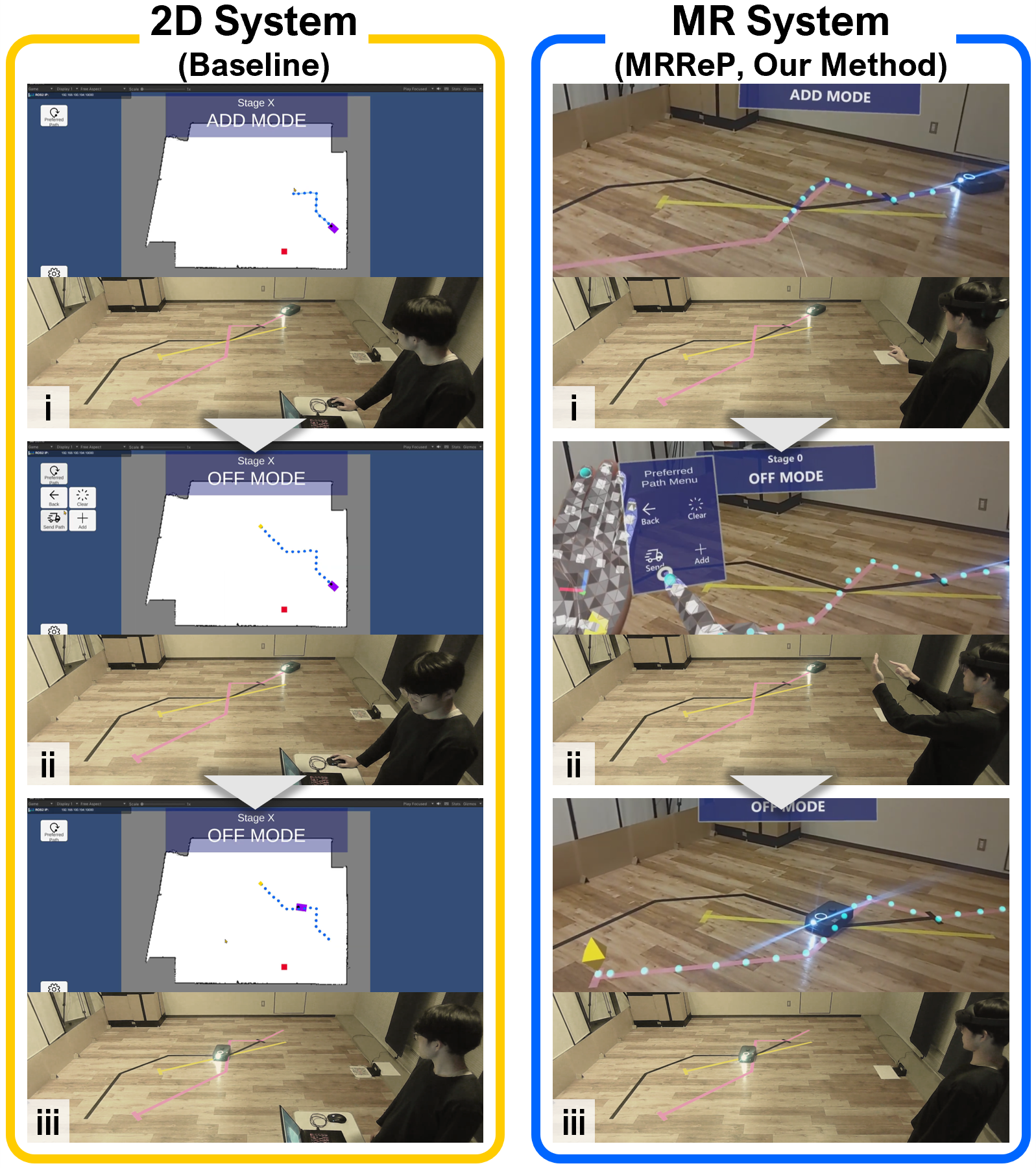}
    \caption{Task procedure. (i) Path drawing. (ii) Path submission. (iii) Robot navigation.}
    \label{fig:experiment_method}
\end{figure}

We conducted a within-subject experiment to evaluate MRReP against a conventional 2D baseline interface.

\subsection{Experimental Method}

Participants used MRReP and a 2D baseline system to draw a Hand-drawn Reference Path (HRP) corresponding to a predefined floor trajectory. After submission, the HRP was transmitted to ROS~2, and the robot autonomously followed the specified path.

The evaluation metrics were as follows:
\begin{itemize}
    \item \textbf{Hand-drawn Path Accuracy}: Spatial deviation between the target path and the participant's HRP.
    \item \textbf{Number of Drawing Attempts}: Total number of drawing operations.
    \item \textbf{Task Completion Time}: Time from the start cue to path submission.
    \item \textbf{Path Stability}: Qualitative comparison of the generated paths and robot trajectories across participants and conditions.
    \item \textbf{Usability and Cognitive Load}: Subjective ratings obtained from questionnaires.
\end{itemize}

For comparison, we implemented a 2D baseline system. MRReP ran on HoloLens~2 and enabled participants to draw paths on the floor using hand gestures, whereas the baseline ran on a laptop PC and enabled path drawing on a 2D map using a mouse. Both systems provided the same three functions (ADD, CLEAR, and SEND) and shared the same navigation pipeline after path submission.

\subsection{Experimental Setup and Procedure}

Fig.~\ref{fig:experiment_environment} shows the experimental environment, which consisted of a practice stage and two test stages (Stages A and B). In each stage, the target path was indicated by tape on the floor, as shown in Fig.~\ref{fig:experimental_environment_gt_map}.
Let $\Sigma_{\mathrm{M}}$ denote the map coordinate system.
The practice path was marked with pink tape, Stage A with yellow tape, and Stage B with black tape. Stage A used a straight path, whereas Stage B used a piecewise linear path with multiple 45-degree turns.

In this experiment, the waypoint generation threshold introduced in Section~\ref{subsec:addition_and_clearing} was fixed at $D_{\mathrm{th}} = \SI{0.2}{\metre}$.

Sixteen participants (10 male and 6 female, aged 19--28 years) took part in the experiment. To reduce order effects, they were evenly divided into two groups, and the order of system use was counterbalanced.

Each experiment lasted approximately 80~min, including an introduction and informed consent (10~min), two system sessions (30~min each), a 5-min break between the sessions, and a final comparative questionnaire (5~min). After the introduction, each participant completed two sessions, one with MRReP and the other with the baseline system. Each session consisted of four phases: instruction, practice, main task, and a post-session questionnaire.

As shown in Fig.~\ref{fig:experiment_method}, in the main task, participants drew an HRP corresponding to the taped floor trajectory after the experimenter's start cue. They then pressed the ``Send'' button, which transmitted the path to ROS~2 and triggered robot navigation. After the first session, participants took a 5-min break and then repeated the same procedure with the other system. After completing both sessions, they answered a final comparative questionnaire.

\begin{table}[t]
    \centering
    \caption{Percentage of HRP within the GT.}
    \label{tab:percentage_within_gt}
    \begin{tabular}{cccc}
        \toprule
        & \multicolumn{2}{c}{Median [IQR]} & $p$-value \\
        \cmidrule(lr){2-3}
        & \: 2D & MR \: & \\
        \midrule
        Stage A & \: 97.0\% [12.4] & 100\% [0.137] \: & 0.0712 \\
        Stage B & \: 65.7\% [19.4] & 100\% [6.17] \: & 0.00121 \\
        \bottomrule
    \end{tabular}
\end{table}

\begin{figure}[t]
    \centering
    \begin{minipage}[b]{0.4\linewidth}
         \centering
         \includegraphics[keepaspectratio, width=0.8\linewidth]{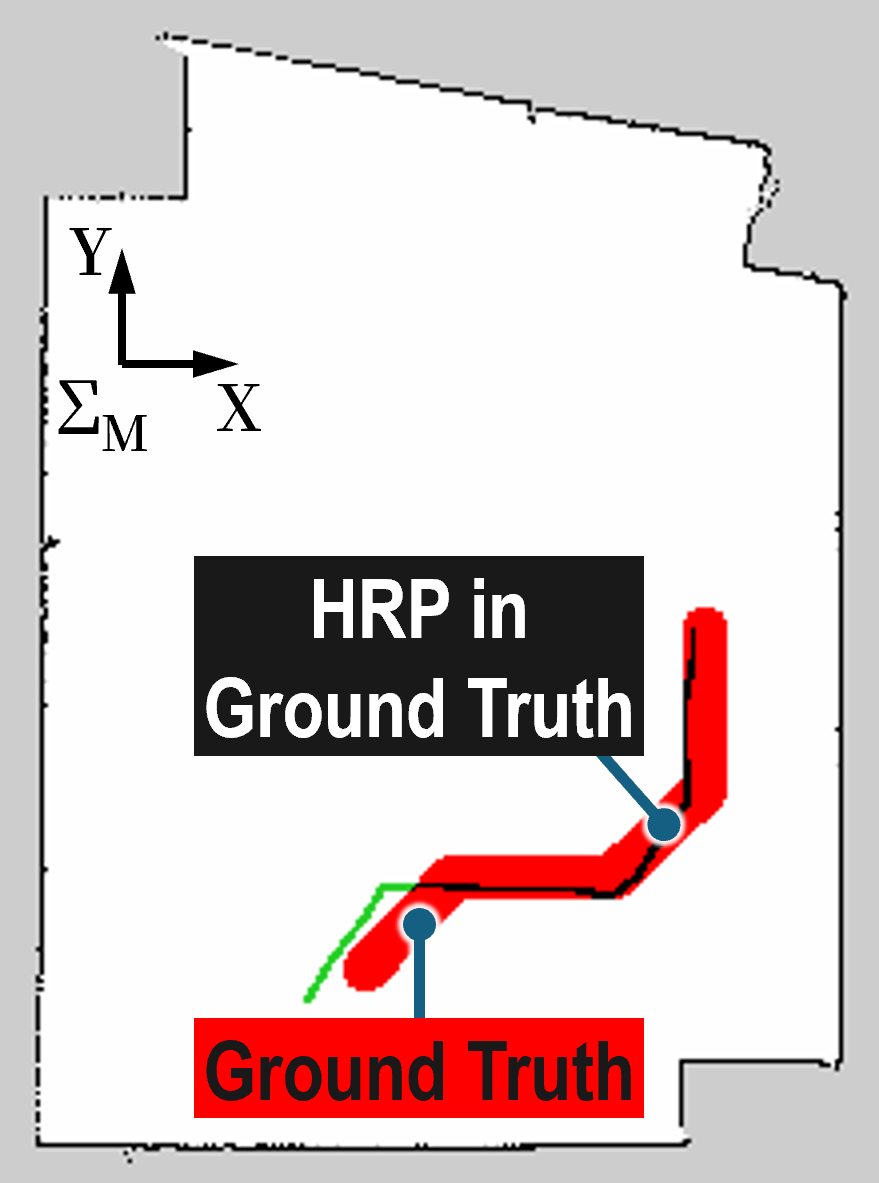}
         \subcaption{HRP and GT.}
         \label{fig:hrp_within_gt}
    \end{minipage}
    \begin{minipage}[b]{0.58\linewidth}
        \centering
        \includegraphics[keepaspectratio, width=0.8\linewidth]{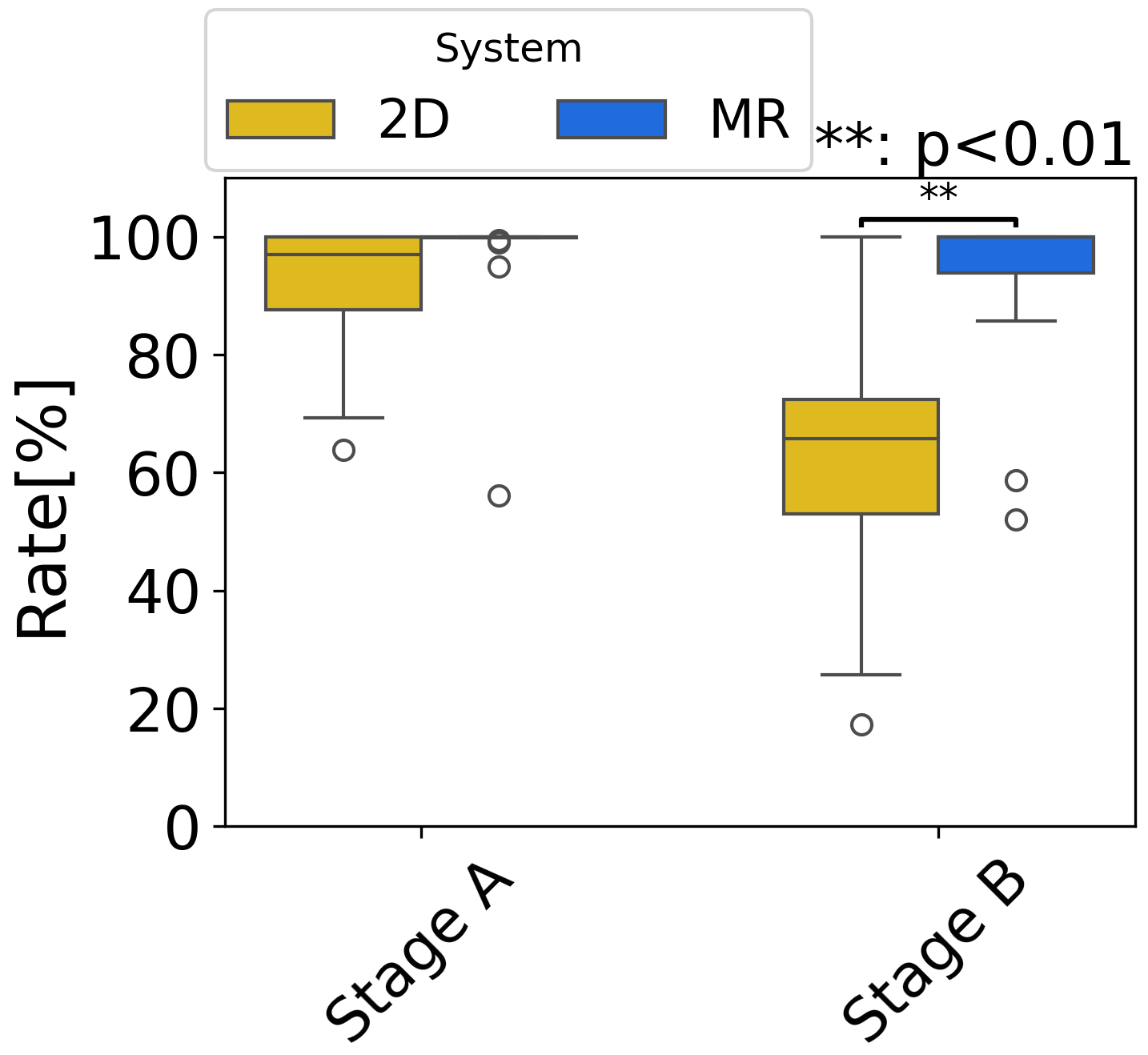}
        \subcaption{Percentage of HRP within the GT.}
        \label{fig:percentage_within_gt_boxplot}
    \end{minipage}
    \caption{Evaluation of HRP coverage within the GT.}
    \label{fig:percentage_within_gt}
\end{figure}

\begin{figure}[t]
    \centering
    \includegraphics[keepaspectratio, width=0.97\linewidth]{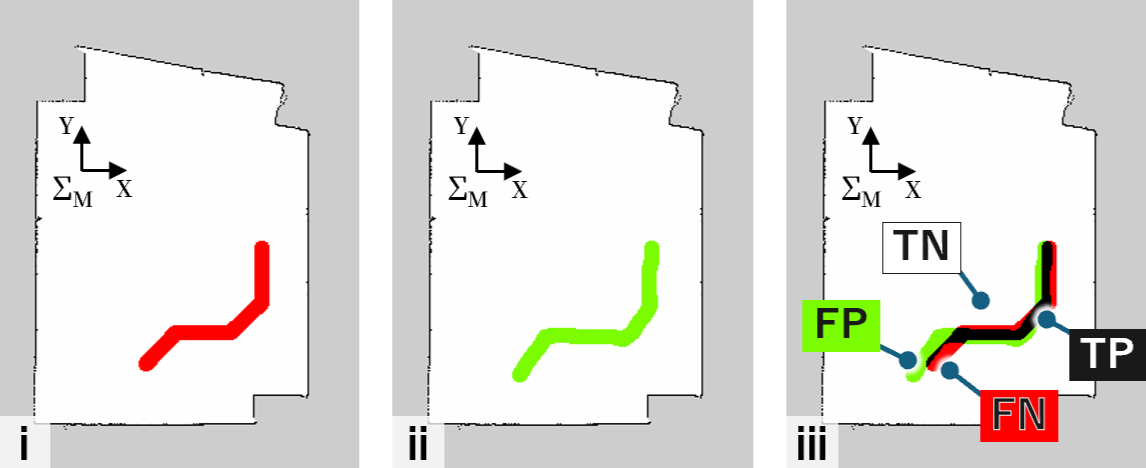}
    \caption{Pixel-wise comparison between the GT and drawn region. (i) Ground Truth. (ii) Example of drawn region on the grid map. (iii) Corresponding pixel classification. TP is shown in black, FP in green, FN in red, and TN in white.}
    \label{fig:path_map_comparing}
\end{figure}

\begin{figure}[t]
    \centering
    \begin{minipage}[b]{0.49\linewidth}
         \centering
         \includegraphics[keepaspectratio, width=1\linewidth]{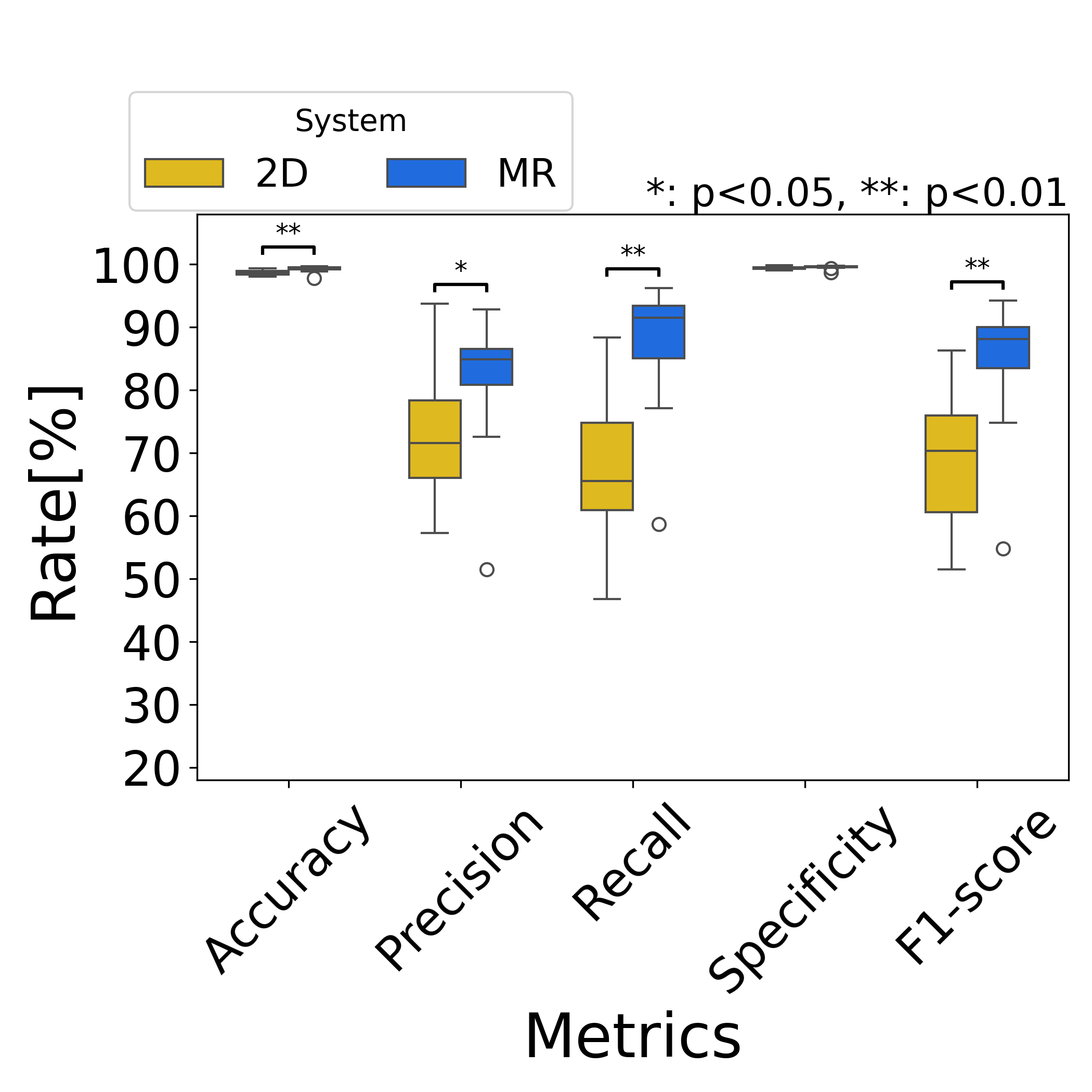}
         \subcaption{Stage A}
         \label{fig:path_difference_index_A_boxplot}
    \end{minipage}
    \begin{minipage}[b]{0.49\linewidth}
         \centering
         \includegraphics[keepaspectratio, width=1\linewidth]{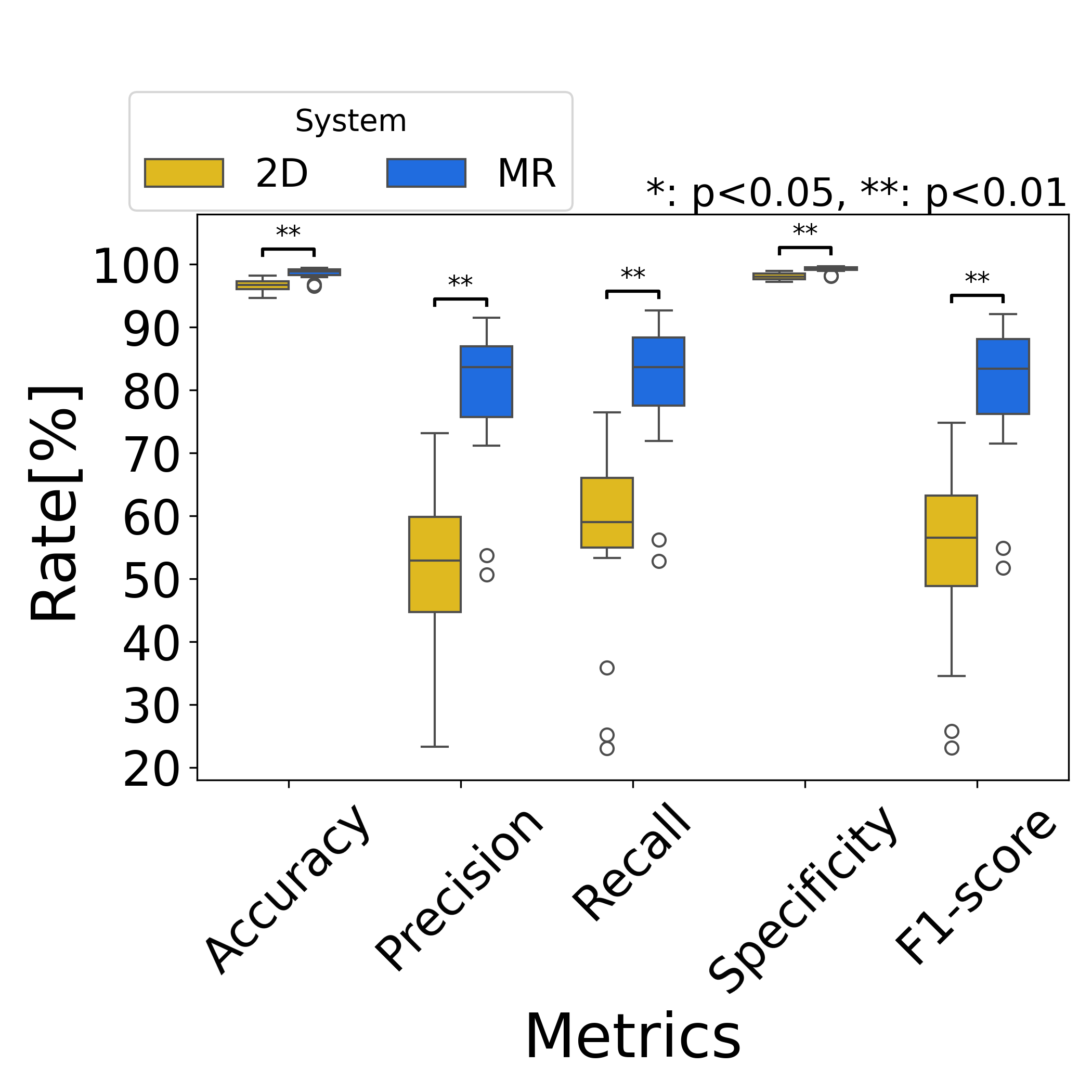}
         \subcaption{Stage B}
         \label{fig:path_difference_index_B_boxplot}
    \end{minipage}
    \caption{HRP accuracy by stage.}
    \label{fig:path_difference_index_boxplot}
\end{figure}

\begin{table}[t]
    \centering
    \caption{HRP accuracy metrics.}
    \label{tab:path_difference}
    \scalebox{0.7}{
    \begin{tabular}{cccccccccccc}
        \toprule
        \multirow{2}{*}{\begin{tabular}{c}Stage\end{tabular}} & & \multicolumn{2}{c}{Accuracy} & \multicolumn{2}{c}{Precision} & \multicolumn{2}{c}{Recall} & \multicolumn{2}{c}{Specificity} & \multicolumn{2}{c}{F1-Score} \\
        \cmidrule(lr){3-4} \cmidrule(lr){5-6} \cmidrule(lr){7-8} \cmidrule(lr){9-10} \cmidrule(lr){11-12}
        &  & \: 2D & MR \: & \: 2D & MR \: & \: 2D & MR \: & \: 2D & MR \: & \: 2D & MR \: \\
        \midrule\midrule
        \multirow{2}{*}{A} & $M$\textsuperscript{\dag}[\%] & 98.7 & 99.4 & 71.6 & 84.9 & 65.5 & 91.6 & 99.4 & 99.6 & 70.3 & 88.1 \\
        & $p$ & \multicolumn{2}{c}{$2.14\times10^{-3}$} & \multicolumn{2}{c}{0.0335} & \multicolumn{2}{c}{$6.10\times10^{-5}$} & \multicolumn{2}{c}{0.105} & \multicolumn{2}{c}{$4.27\times10^{-4}$} \\
        \midrule
        \multirow{2}{*}{B} & $M$[\%] & 96.8 & 98.9 & 52.9 & 83.6 & 59.0 & 83.7 & 98.1 & 99.4 & 56.5 & 83.4 \\
        & $p$ & \multicolumn{2}{c}{$9.16\times10^{-5}$} & \multicolumn{2}{c}{$9.16\times10^{-5}$} & \multicolumn{2}{c}{$9.16\times10^{-5}$} & \multicolumn{2}{c}{$9.16\times10^{-5}$} & \multicolumn{2}{c}{$9.16\times10^{-5}$} \\
        \bottomrule
        \multicolumn{12}{l}{\textsuperscript{\dag}$M$: Median. Reported values are medians across participants.}
    \end{tabular}
    }
\end{table}

\subsection{Experimental Results}\label{sec:EXP_result}
For each evaluation metric, a Wilcoxon signed-rank test~\cite{rey2011wilcoxon} was used to compare the MR and 2D conditions.
The detailed results for each metric are presented below.

\subsubsection{Hand-drawn Path Accuracy}
The spatial accuracy of the HRP was evaluated by comparing it with the Ground Truth (GT) derived from the target path indicated by tape on the floor.

The GT was defined as the region that serves as the robot's center when the robot's body overlaps with the tape on the floor.
Fig.~\ref{fig:percentage_within_gt}(a) shows an example of the GT and HRP overlaid on the map. In the figure, the GT is shown in red, the HRP in green, and their overlap in black. We first computed the proportion of the HRP lying within the GT. As shown in Table~\ref{tab:percentage_within_gt} and Fig.~\ref{fig:percentage_within_gt}(b), the HRPs drawn with the proposed MR system were largely contained within the GT, with low inter-participant variability.

Next, the HRP was expanded to the same width as the GT, and a pixel-wise comparison between the resulting drawn region and the GT was performed on the robot's grid map. As illustrated in Fig.~\ref{fig:path_map_comparing}, each cell was classified as follows:

\begin{itemize}
    \item \textbf{True Positive (TP):} Cells belonging to both the GT and the drawn region.
    \item \textbf{False Positive (FP):} Cells belonging to the drawn region but not the GT.
    \item \textbf{False Negative (FN):} Cells belonging to the GT but not the drawn region.
    \item \textbf{True Negative (TN):} Cells belonging to neither the GT nor the drawn region.
\end{itemize}

From these classifications, we calculated Accuracy, Precision, Recall, Specificity, and F1-score:

\begin{itemize}
    \item \textbf{Accuracy:} Proportion of correctly classified cells over the entire grid map.
    \item \textbf{Precision:} Proportion of drawn-region cells that fall within the GT.
    \item \textbf{Recall:} Proportion of GT cells covered by the drawn region.
    \item \textbf{Specificity:} Proportion of non-GT cells where no path was drawn.
    \item \textbf{F1-score:} Harmonic mean of Precision and Recall.
\end{itemize}

Because the purpose of this analysis was to evaluate how faithfully participants reproduced the target path itself, we focused primarily on Precision, Recall, and F1-score. Precision captures the extent to which the drawn region stayed within the GT without unnecessary protrusion into non-target areas, whereas Recall captures how completely the drawn region covered the GT without omissions or unintended shortcuts. F1-score was considered as a balanced summary of these two aspects. In contrast, Accuracy and Specificity are strongly influenced by the large number of background cells outside the path region and therefore are less informative for assessing the fidelity of path reproduction. 
Table~\ref{tab:path_difference} and Fig.~\ref{fig:path_difference_index_boxplot} summarize the results.

\begin{figure}[t]
    \centering
    \begin{minipage}[b]{0.46\linewidth}
         \centering
         \includegraphics[keepaspectratio, width=0.8\linewidth]{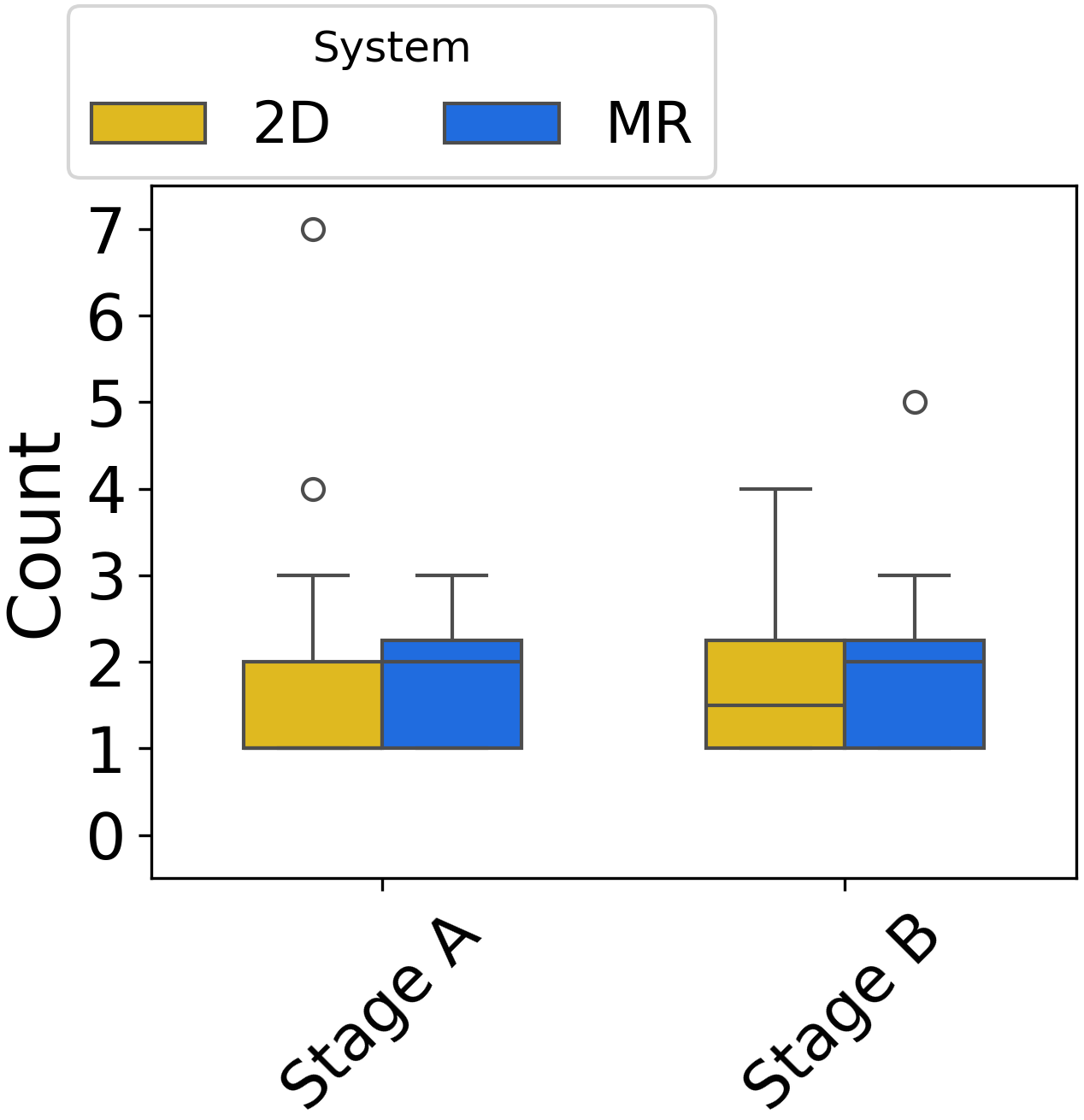}
         \subcaption{Drawing operations}
         \label{fig:drawing_number_boxplot}
    \end{minipage}
    \begin{minipage}[b]{0.52\linewidth}
         \centering
         \includegraphics[keepaspectratio, width=0.8\linewidth]{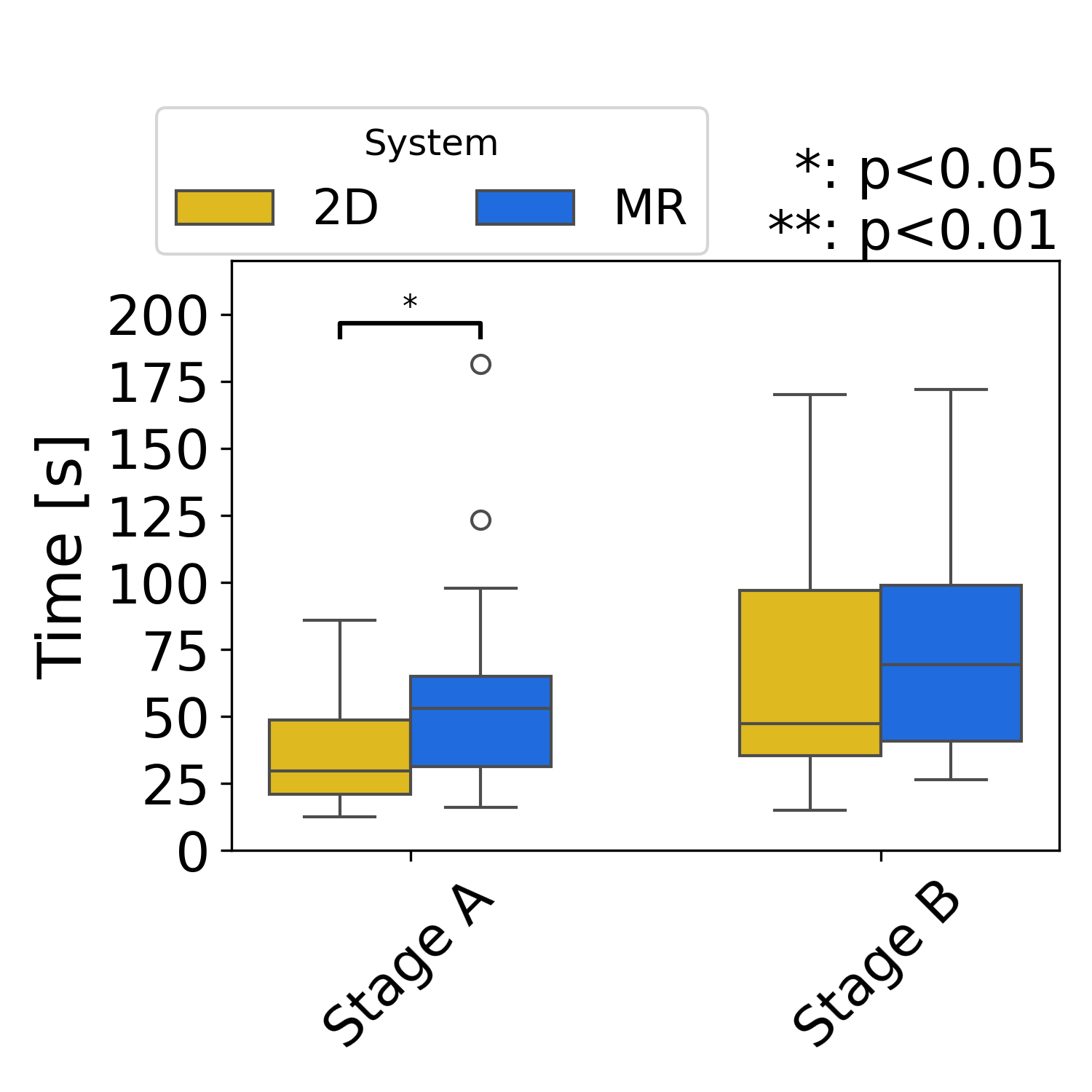}
         \subcaption{Task completion time}
         \label{fig:task_time_boxplot}
    \end{minipage}
    \caption{Drawing operations and task completion time.}
    \label{fig:drawing_number_task_time_boxplot}
\end{figure}

\begin{table}[t]
    \centering
    \caption{Task completion time [sec].}
    \label{tab:task_completion_time}
    \begin{tabular}{cccc}
        \toprule
        & \multicolumn{2}{c}{Median [IQR]} & $p$-value \\
        \cmidrule(lr){2-3}
        & \: 2D & MR \: & \\
        \midrule
        Stage A & \: 29.6 [27.9] & 53.0 [33.4] \: & 0.0290 \\
        Stage B & \: 47.2 [61.6] & 69.3 [58.1] \: & 0.323 \\
        \bottomrule
    \end{tabular}
\end{table}

In both Stage A and Stage B, the MR system achieved higher scores on the primary fidelity-related metrics.
For Precision, the MR system significantly outperformed the 2D system, 
achieving 84.9\% vs.\ 71.6\% in Stage A and 83.6\% vs.\ 52.9\% in Stage B.
These results indicate that the MR system enabled more accurate alignment with the GT while suppressing protrusion into non-target areas. Inter-participant variability was also smaller in the MR condition, particularly in Stage B.

For Recall, the MR system again significantly outperformed the 2D system, 
with 91.6\% vs.\ 65.5\% in Stage A and 83.7\% vs.\ 59.0\% in Stage B.
This suggests that the MR system allowed participants to cover the target trajectory more completely, reducing omissions and unintended shortcuts.

For the remaining metrics, the MR system also showed higher values than the 2D system, except for Specificity in Stage A, and most differences were statistically significant.

Overall, these results indicate that the proposed MR system enabled participants to reproduce the intended trajectory more faithfully, thereby allowing the robot to follow a route closer to the operator's intentions.

\subsubsection{Number of Drawing Operations}

Fig.~\ref{fig:drawing_number_task_time_boxplot}(a) shows the number of path drawing attempts for each system. Because only one path had to be drawn in each stage, a value closer to 1 is preferable. No statistically significant difference was observed between the two systems.

The absence of a significant difference in drawing attempts suggests different causes of repeated adjustments in the two systems. In MRReP, hand jitter during gesture input and the direct visibility of small deviations between the GT and HRP may have encouraged repeated fine adjustment. In the 2D system, although mouse-based input was stable, mental translation between the 2D map and the physical environment may also have caused re-drawing. These factors may explain why the number of drawing attempts did not differ significantly between the two systems.

\subsubsection{Task Completion Time}
Task completion time was measured from the experimenter's start cue until the participant finished drawing the path and pressed the ``Send'' button.

As shown in Table~\ref{tab:task_completion_time} and Fig.~\ref{fig:drawing_number_task_time_boxplot}(b), the MR system tended to require more time than the 2D system. As in the previous subsection, this may reflect both the need for finer adjustment in MRReP and the larger physical movements required for gesture-based input than for mouse-based operation.

\subsubsection{Path Stability}
\begin{figure}[t]
    \centering
    \begin{minipage}{\linewidth}
        \centering
        \includegraphics[width=\linewidth]{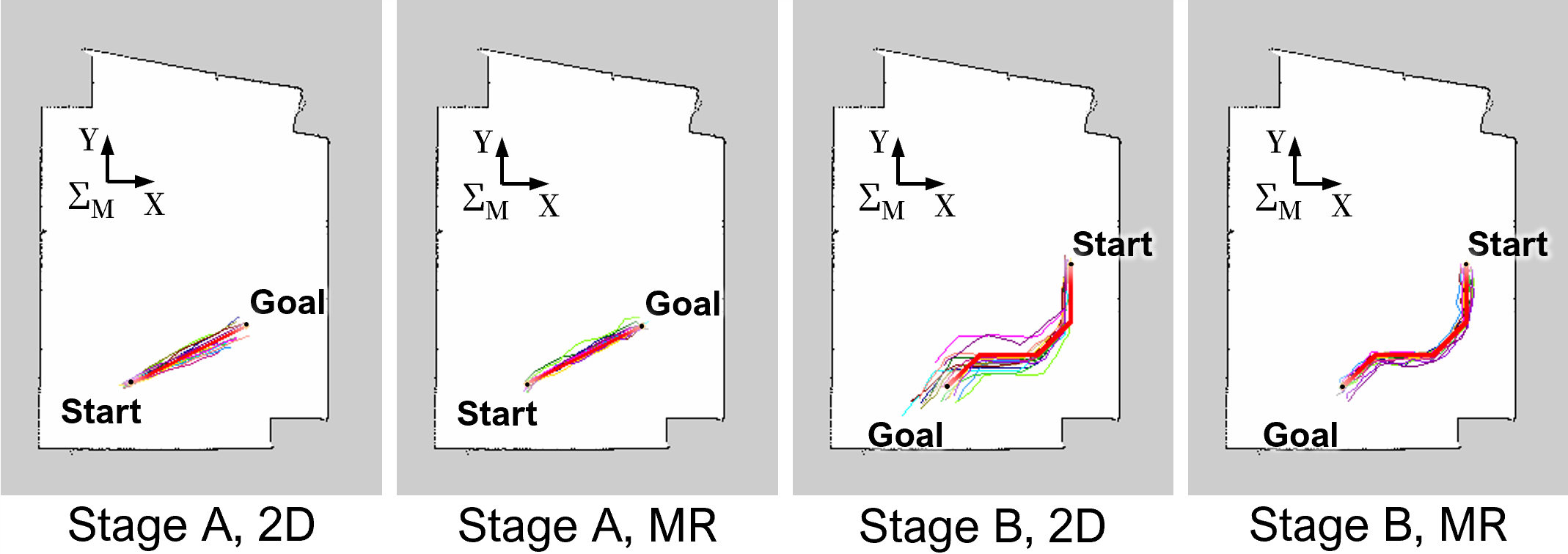}
        \subcaption{Global paths}
    \end{minipage}    

    \vspace{0.5em}

    \begin{minipage}{\linewidth}
        \centering
        \includegraphics[width=\linewidth]{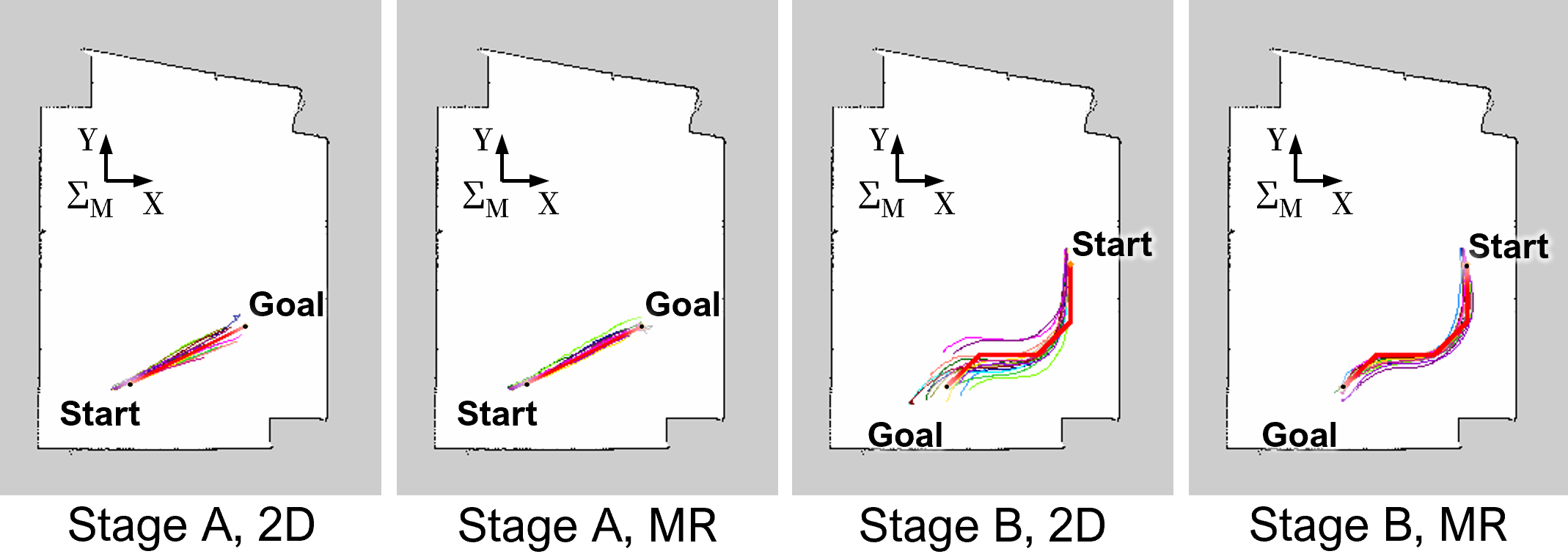}
        \subcaption{Robot trajectories}
    \end{minipage}

    \caption{Global paths and robot trajectories by stage and condition.}
    \label{fig:global_paths_robot_trajectories_map}
\end{figure}

We evaluated the planned global paths and actual robot trajectories derived from the HRPs. Fig.~\ref{fig:global_paths_robot_trajectories_map} shows the global paths and robot trajectories for both systems in Stages A and B. The GT is shown as a solid red line, with the HRPs of all participants superimposed.

In the 2D condition, paths in Stage A often terminated before reaching the GT endpoint. In both stages, deviation from the GT tended to increase with distance from the start point, likely due to the limited depth perception of a 2D screen.

In contrast, the MR system showed smaller deviations from the GT and lower inter-subject variance in both stages. These results suggest that the MR system enabled more stable and accurate path specification, even for the more complex trajectory.

\subsubsection{Usability and Cognitive Load}

\begin{table}[t]
    \centering
    \caption{Subjective evaluation results.}
    \label{tab:sus_nasa_overall}
    \begin{tabular}{lcccc}
        \toprule
        Metric & \multicolumn{2}{c}{Median [IQR]} & Effect size & $p$-value \\
        \cmidrule(lr){2-3}
        & \: 2D & MR \: & & \\
        \midrule
        SUS      & 51.3 [23.1] & 75.0 [16.3] & 0.711 & 0.00586\\
        NASA-TLX & 61.5 [22.1]  & 47.7 [21.3] & 0.621 & 0.0110\\
        \bottomrule
    \end{tabular}
\end{table}

\begin{figure}[t]
    \centering
    \begin{minipage}[b]{0.45\linewidth}
        \centering
        \includegraphics[keepaspectratio, width=1\linewidth]{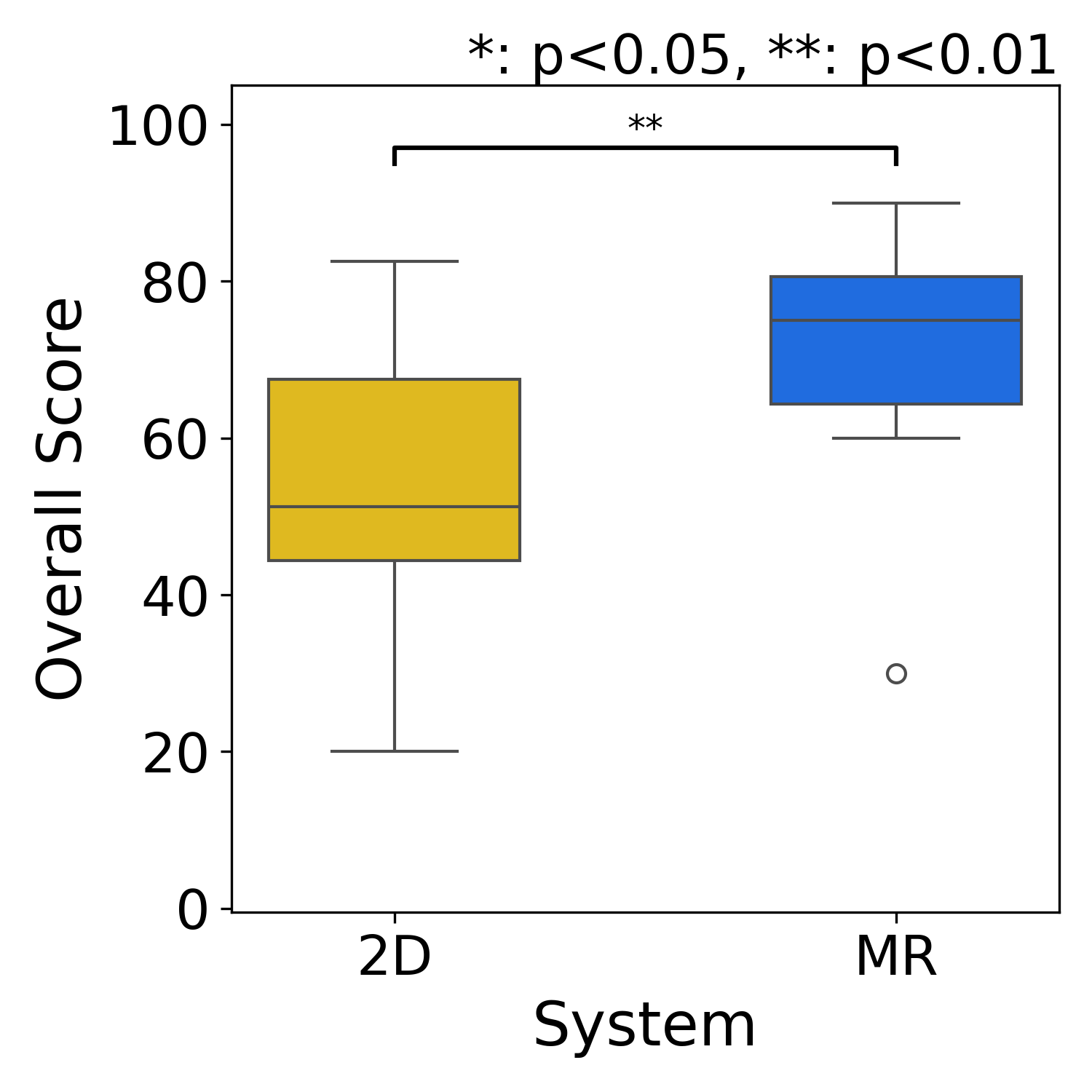}
        \subcaption{SUS}
        \label{fig:sus_boxplot}
    \end{minipage}
    \begin{minipage}[b]{0.45\linewidth}
        \centering
        \includegraphics[keepaspectratio, width=1\linewidth]{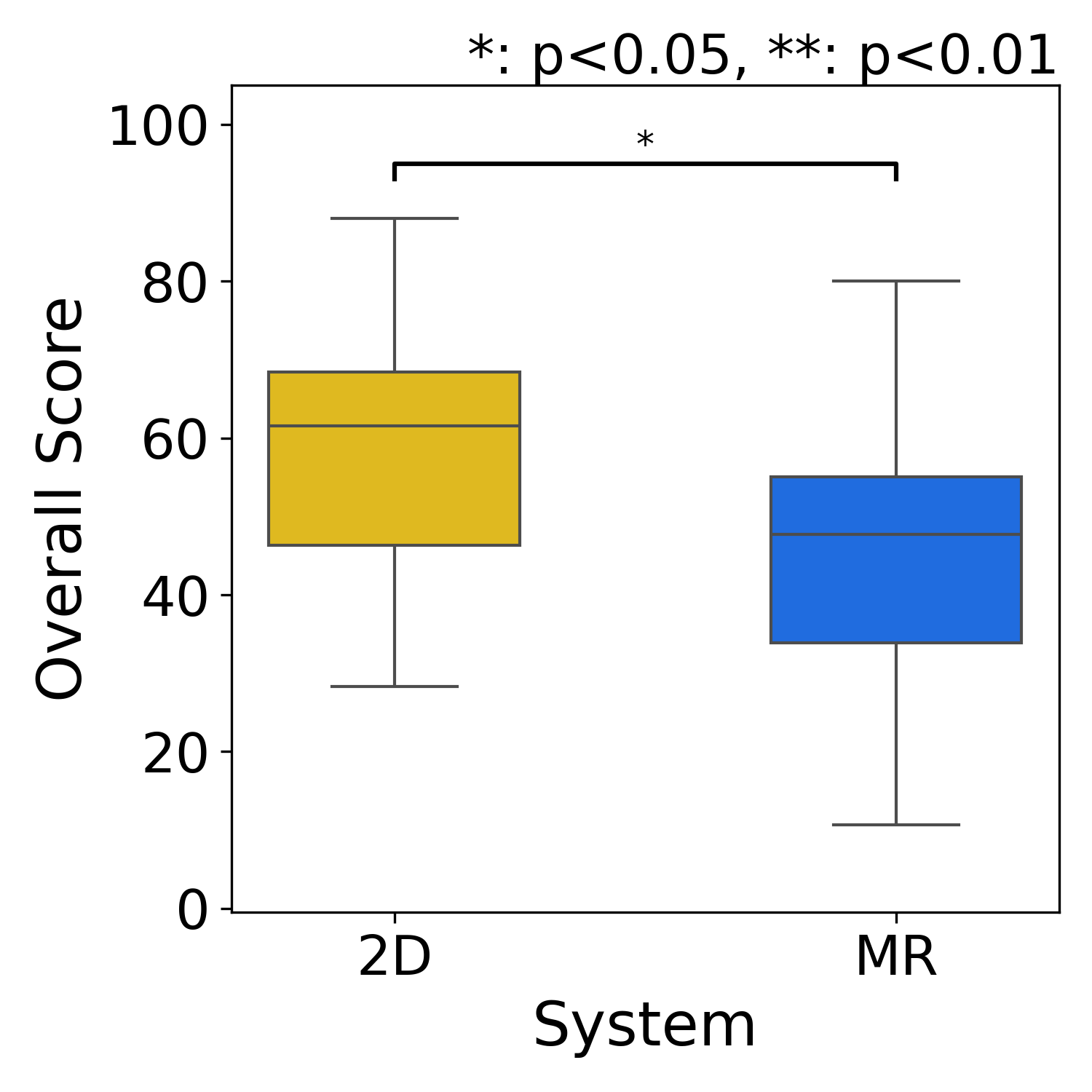}
        \subcaption{NASA-TLX}
        \label{fig:nasa_tlx_boxplot}
    \end{minipage}
    \caption{Subjective evaluation scores.}
    \label{fig:sus_nasa_overall_boxplot}
\end{figure}

Usability and mental workload were evaluated using the System Usability Scale (SUS)~\cite{brooke1996sus} and the NASA Task Load Index (NASA-TLX)~\cite{hart1988development}, respectively.

Effect sizes were also calculated to quantify the magnitude of the differences.
Table~\ref{tab:sus_nasa_overall} summarizes the medians, interquartile ranges, and effect sizes.

As shown in Table~\ref{tab:sus_nasa_overall} and Fig.~\ref{fig:sus_nasa_overall_boxplot}, the MR system achieved higher SUS scores and lower NASA-TLX scores than the 2D system, indicating better usability and lower workload.

However, several MR-specific challenges were identified, including physical discomfort such as arm fatigue during gesture-based interaction with HoloLens~2 and stress caused by gesture misrecognition, which may reflect the larger physical movements required than in the 2D system and participants' limited familiarity with MR operation.

\section{Conclusions}\label{sec:CON}
This paper presented MRReP, a Mixed Reality-based interface that enables users to specify robot navigation paths through hand gestures in the physical environment. By directly connecting a user-drawn HRP with the robot navigation stack, the proposed system supports more faithful path specification than conventional 2D interfaces.

A within-subject experiment showed that MRReP improved path specification accuracy, usability, and perceived workload relative to the 2D baseline. These results suggest that direct path specification in the physical environment using MR is an effective approach for incorporating human spatial intention into mobile robot navigation.

\bibliographystyle{IEEEtran}

\end{document}